%% file: main.tex
\documentclass[11pt,a4paper]{article}
\usepackage[hyperref]{emnlp2020}
\usepackage{times}
\usepackage{latexsym}
\usepackage{url}
\usepackage{todonotes}
\usepackage{array}
\usepackage{amsmath}
\usepackage{graphicx}
\usepackage{bbm}
\usepackage{natbib}
\usepackage{svg}
\graphicspath{{src/img/}}
\usepackage{multirow}
\usepackage{multicol}
\usepackage{booktabs}
\usepackage{appendix}
\usepackage{bm}
\usepackage{amssymb}
\usepackage{lipsum}
\newcommand{\start}[1]{\smallskip\vspace{0.5mm}\noindent{{\bf #1}}}

\input{macro}
\input{math_commands}

\usepackage{microtype}

\aclfinalcopy 
\def\aclpaperid{1466} 

\title{Learning Collaborative Agents with Rule Guidance for\\ Knowledge Graph Reasoning}

\author{Deren Lei$^{1}$\thanks{\enspace Equal contributions.}, Gangrong Jiang$^{1}$\footnotemark[1], Xiaotao Gu$^2$, Kexuan Sun$^{1}$, Yuning Mao$^{2}$, Xiang Ren$^1$\\
$^1$University of Southern California\\
$^2$University of Illinois at Urbana-Champaign \\
\{derenlei, gjiang, kexuansu, xiangren\}@usc.edu, \{xiaotao2, yuningm2\}@illinois.edu\\
}

\begin{document}
\maketitle
\input{abs}

\input{src/1-Intro.tex}

\input{src/2-Prelim.tex}
\input{src/3-Method.tex}

\input{src/4-Exp.tex}
\input{src/6-Conclusion.tex}

\bibliography{main}
\bibliographystyle{acl_natbib}

\input{appendix.tex}

\end{document}

%% file: macro.tex
\newcommand{\mE}{\mathcal{E}}

\newcommand{\mR}{\mathcal{R}}

\newcommand{\highest}[1]{\textbf{#1}}



%% file: math_commands.tex
\usepackage{amsmath,amsfonts,bm}

\def\eqref#1{equation~\ref{#1}}

\def\1{\bm{1}}

\def\vd{{\bm{d}}}
\def\ve{{\bm{e}}}

\def\vh{{\bm{h}}}

\def\vr{{\bm{r}}}

\def\mE{{\bm{E}}}

\def\mR{{\bm{R}}}

\def\mW{{\bm{W}}}

\DeclareMathAlphabet{\mathsfit}{\encodingdefault}{\sfdefault}{m}{sl}
\SetMathAlphabet{\mathsfit}{bold}{\encodingdefault}{\sfdefault}{bx}{n}



%% file: abs.tex
\begin{abstract}
Walk-based models have shown their advantages in knowledge graph (KG) reasoning by achieving decent performance while providing interpretable decisions.
However, the sparse reward signals offered by the KG during traversal are often insufficient to guide a sophisticated walk-based reinforcement learning (RL) model.
An alternate approach is to use traditional symbolic methods (e.g., rule induction), which achieve good performance but can be hard to generalize due to the limitation of symbolic representation.
In this paper, we propose \textit{RuleGuider}, which leverages high-quality rules generated by symbolic-based methods to provide reward supervision for walk-based agents. Experiments on benchmark datasets show that \textit{RuleGuider} improves the performance of walk-based models without losing interpretability. 
\footnote{\small \url{https://github.com/derenlei/KG-RuleGuider}}
\end{abstract}

%% file: src/1-Intro.tex
\section{Introduction}



While knowledge graphs (KGs) are widely adopted in natural language processing applications, a major bottleneck hindering its usage is the sparsity of 
facts~\cite{min2013distant}, leading to extensive studies on KG completion (or reasoning) \cite{Trouillon2016ComplexEF,dettmers2018convolutional,das2017go,xiong2017deeppath,lin-etal-2018-multi-hop,Meilicke2019AnytimeBR}. 
Many traditional approaches on the KG reasoning task are based on logic rules \cite{landwehr2007integrating,landwehr2010fast, galarraga2013amie,galarraga2015fast}. 
These methods are referred to as \textit{symbolic-based methods}. Although they showed good performance~\cite{Meilicke2019AnytimeBR,meilicke2020reinforced}, they are inherently limited by their representations and generalizability of the associated relations of the given rules.
To ameliorate such limitations, \textit{embedding-based methods}~\cite{Bordes2013TranslatingEF,Socher2013ReasoningWN,wang2014knowledge,Yang2014EmbeddingEA,Trouillon2016ComplexEF,dettmers2018convolutional, Dettmers2017Convolutional2K,Sun2019RotatEKG,Zhang_2019} were proposed. They learn distributed representations for entities and relations and make predictions using the representations. Despite their superior performance, 
they fail to make human-friendly interpretations.


To improve the interpretability,
many recent efforts formulate the task as a multi-hop reasoning problem using reinforcement learning (RL) techniques~\cite{xiong2017deeppath, das2017go, shen2018reinforcewalk,chen2018variational,lin-etal-2018-multi-hop}, referred to as \textit{walk-based methods}.
A major issue of these methods is the reward function. A ``hit or not'' reward is too sparse while a shaped reward using an embedding-based distance measurement~\citet{lin-etal-2018-multi-hop} may not always result in desirable paths.


In this paper, 
we propose \textit{RuleGuider} to
tackle the aforementioned reward issue in walk-based methods with the help of symbolic rules. We aim to improve the performance of walk-based methods without losing their interpretability. The RuleGuider is composed of a symbolic-based model fetching logic rules and a walk-based agent searching reasoning paths with the guidance of the rules.
We also introduce a way to separate the walk-based agent to allow for further efficiency.
We experimentally show the efficiency of our model without losing the interpretability.





%% file: src/2-Prelim.tex
\section{Problem and Preliminaries}
\label{sec:problem}
In this section, we review the KG reasoning task. 
We also describe the symbolic-based and walk-based methods used in RuleGuider.

\start{Problem Formulation.}
A KG consisting of fact triples is represented as $\mathcal{G} = \{(e_i, r, e_j)\} \subseteq \mathcal{E} \times \mathcal{R} \times \mathcal{E}$, where $\mathcal{E}$ and $\mathcal{R}$ are the set of entities and relations, respectively.
Given a query $(e_s, r_q, ?)$ where $e_s$ is a subject entity and $r_q$ is a query relation, the task of KG reasoning is to find a set of object entities $E_o$ such that $(e_s, r_q, e_o)$ , where $e_o\in E_o$, is a fact triple missing in $\mathcal{G}$.
We denote the queries $(e_s, r_q, ?)$ as \textit{tail queries}. We note that we can also perform \textit{head queries} $(?, r_q, e_o)$. To be consistent with most existing works, we only consider tail queries in this paper.


\begin{figure}[t]
    \centering
    \includegraphics[width=7.5cm]{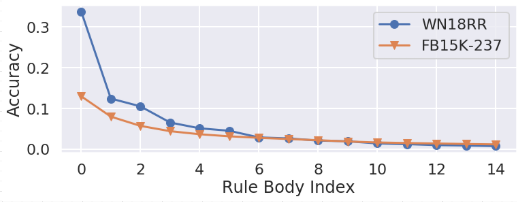}
    \vspace{-.2cm}
    \caption{
    Rule quality difference between datasets. There are exists high quality rules on WN18RR.}
    \vspace{-.25cm}
    \label{fig:percentage}
\end{figure}

\start{Symbolic-based Methods.}
Some previous methods mine Horn rules from the KG and predict missing facts by grounding these rules.
A recent method AnyBURL \cite{Meilicke2019AnytimeBR} showed comparable performance to the state-of-the-art embedding-based methods.
It first mines rules by sampling paths from the $\mathcal{G}$, and then make predictions by matching queries to the rules. Rules are in the format: $ r(X, Y) \leftarrow b_1(X, A_2) \land...\land b_n(A_n, Y)$, where
upper-case letters represent variables.
A \emph{rule head} is denoted by $r(\cdots)$ and a \emph{rule body} is denoted by the conjunction of atoms $b_1(\cdots), \dots, b_n(\cdots)$. We note that $r(c_i, c_j)$ is equivalent to the fact triple $(c_i, r, c_j)$.

However, these methods have limitations. 
For example, rules mined from different KGs may have different qualities, which makes the reasoner hard to select rules. Figure~\ref{fig:percentage} shows such difference. Rules are sorted based on accuracy of predicting the target entities. The top rules from WN18RR are much more valuable than those from FB15K-237.

\start{Walk-based Methods.}
Given a query $(e_s, r_q, ?)$,
walk-based methods train an RL agent to find a path from $e_s$ to the desired object entity $e_o$ that implies the query relation $r_q$.
At step $t$, the current state is represented by a tuple $s_t = (e_t, (e_s, r_q))$, where $e_t$ is the current entity. 
The agent then samples the next relation-entity pair to visit from possible actions $A_t = \{(r', e') | (e_t, r', e') \in \mathcal{G}\}$. The agent receives a reward when it reaches $e_o$.

%% file: src/3-Method.tex
\section{Proposed Method: RuleGuider}
\label{Proposed}

\begin{figure}
    \centering
    \includegraphics[width=8cm]{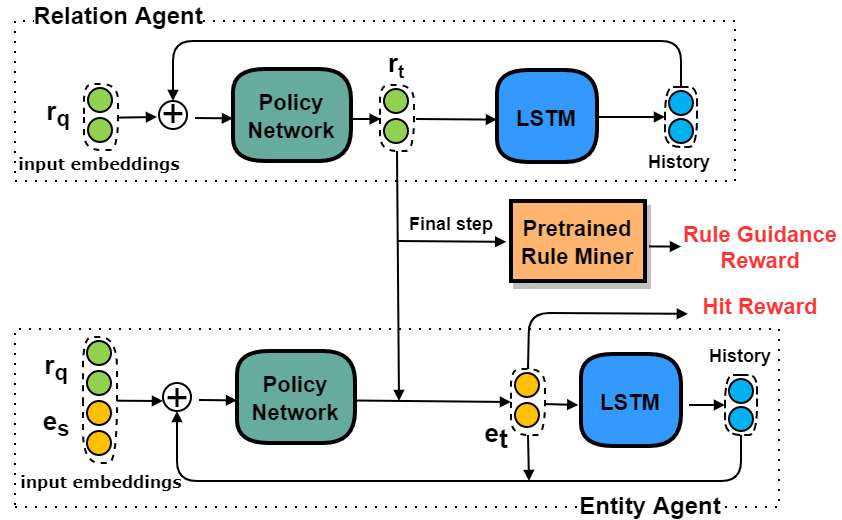}
    \vspace{-.25cm}
    \caption{\textbf{The architecture of two agents}. The relation and entity agent interact with each other to generate a path. At each step, the entity agent first selects an entity from valid entities.
    The relation agent then samples a relation based on the selected entity.
    At the final step, they receive a hit reward based on the last selected entity and a rule guidance reward from the pre-mined rule set based on the selected path.}
    \vspace{-.35cm}
    \label{fig:my_label}
\end{figure}




RuleGuider consists of a symbolic-based method (see Section~\ref{sec:problem}), referred to as \textit{rule miner}, and a walk-based method, referred to as \textit{agent}.
The rule miner first mines logic rules and the agent traverses over the KG to learn the probability distribution of reasoning paths with the guidance (via the reward) of the rules.
As the agent walks through relations and entities alternatively, we propose to separate the agent into two sub-agents: a relation and entity agents. After the separation, the search space is significantly pruned.
Figure \ref{fig:my_label} shows the structure of these two agents in detail.

\subsection{Model Architecture}

\start{Relation Agent.} 
At step $t$ ($t = 1, \cdots, T$, $T$ is the number of hops), the relation agent selects a single relation $r_t$ which is incident to the current entity $e_{t-1}$, where $e_0$=$e_s$. Given a query $(e_s, r_q, ?)$ and a set of rules \textbf{R}, this process can be formulated as $r_t = P^R(r_q, e_{t-1}, \mathbf{R}, \vh_t^R)$ where $\vh_t^R$ is the relation history. The agent first filter out rules whose heads are not same as $r_q$, and then it selects $r_t$ from the $t^{th}$ atoms of the remaining rule bodies, i.e. $b_t(\cdots)$ in the rule pattern. 

Since the rule miner provides confidence scores of rules, we first use RL techniques to pre-train this agent using the scores.
During training, the agent applies the pre-trained strategy (distribution) and keeps tuning the distribution by utilizing semantic information provided by embeddings. In another words, the relation agent leverages both confidence scores of pre-mined rules as well as embedding shaped hit rewards.


\start{Entity Agent.}
At step $t$, the agent generates the distribution of all candidate entities based on $e_s$, $r_q$, and the entity history $\vh_t^E$. Given the current relation $r_t$, this process can formally be represented as $e_t = P^E(e_s, r_q, r_t, \vh_t^E)$. The agent selects an entity from all entities that incident on $r_t$. In this way, the entity and relation agent can reason independently.

In experiments, we have also tried to let the entity agent generate distribution based on relation agent pruned entity space. In this way, the entity agent takes in the selected relation and can leverage the information from the relation agent. However, the entity space may be extremely small and hard to learn. It makes the entity agent less effective, especially on large and dense KG.


\start{Policy Network.}
The relation agent's search policy is parameterized by the embedding of $\vr_q$ and $\vh_t^R$. The relation history is encoded using an LSTM\cite{hochreiter1997long}: $\vh_t^R = \text{LSTM}(\vh_{t-1}^R, \vr_{t-1})$, where $\vr_{t-1}\in \mathbb{R}^d$ is the embedding of the last relation. We initialize $\vh_0^R= \text{LSTM} (\bm{0}, \vr_s)$, where $\vr_s$ is a special start relation embedding to form an initial relation-entity pair with source entity embedding $\ve_s$.
Relation space embeddings $\mR_t \in \mathbb{R}^{|R_t|\times d}$ consist embeddings of all the relations in relation space $R_t$ at step t. 
Finally, relation agent outputs a probability distribution $\vd^R_t$ and samples a relation from it.
 \( \vd^R_t = \sigma(\mR_t \times \mW_1\ \text{ReLU} (\mW_2 [\vh_t^R; \vr_q])) \nonumber \)
\noindent
where $\sigma$ is the softmax operator, $\mW_1$ and $\mW_2$ is trainable parameters. We design relation agent's history-dependent policy as $\bm{\pi}^R = (\vd_1^R, \vd_2^R, \dots, \vd_T^R)$.

Similarly, entity agent's history-dependent policy is $\bm{\pi}^E = (\vd_1^E, \vd_2^E, \dots, \vd_T^E)$. Entity agent can acquire its embedding of last step $\ve_{t-1}$, entity space embeddings $\mE_t$, its history $\vh_t^E = \text{LSTM}(\vh_{t-1}^E, \ve_{t-1})$,  and the probability distribution of entities $\vd^E_t$ as follows.
$\vd^E_t = \sigma(\mE_t \times \mW_3 \text{ReLU}(\mW_4 [\vh_t^E; \vr_q; \ve_s; \ve_t]))$
where $\mW_3$ and $\mW_4$ is trainable parameters. Note that entity agent uses a different LSTM to encode the entity history.

\subsection{Model Learning}
We train the model by letting the two aforementioned agents to start from specific entities and traverse through the KG in a fixed number of hops. The agents receive rewards at their final step.

\start{Reward Design.}
Given a query, the relation agent prefers paths which direct the way to the correct object entity. Thus, given a relation path, we give reward according to its confidence retrieved from the rule miner, referred to as \textit{rule guidance reward} $R_r$.
We also add a Laplace smoothing $p_c=5$ to the confidence score for the final $R_r$.



In addition to $R_r$, the agent will also receive a \textit{hit reward} $R_h$, which is 1 if the predicted triple $\epsilon=(e_s, r_q, e_T) \in \mathcal{G}$. 
Otherwise, we use the embedding of $\epsilon$ to measure reward as in \citet{lin-etal-2018-multi-hop}. $R_h=\mathbb{I}(\epsilon \in \mathcal{G})+(1-\mathbb{I}(\epsilon \in \mathcal{G})f(\epsilon)$
where $\mathbb{I}(\cdot)$ is an indicator function, $f(\epsilon)$ is a composition function for reward shaping using embeddings. 

\start{Training Procedure.}
 We train the model in four stages. 1) Train relation and entity embeddings using an embedding-based method. 2) Apply a rule miner to retrieve rules and their associated confidence scores.
 3) Pre-train the relation agent by freezing the entity agent and asking the relation agent to sample a path. We only use the rule miner to evaluate the path and compute $R_r$ based on the pre-mined confidence score.
 4) Jointly train the relation and entity agent to leverage the embeddings to compute $R_h$. The final reward $R$ involves $R_r$ and $R_h$ with a constant factor $\lambda$: $R = \lambda R_r + (1 - \lambda) R_h $. The policy networks of two agents are trained using the REINFORCE \cite{williams1992simple} algorithm to maximize $R$.

%% file: src/4-Exp.tex
\begin{table*}[!t]
    \centering
    \small
    \hspace*{-0.32cm}
    \setlength\tabcolsep{3.1pt}
    \setlength\extrarowheight{3pt}
    \scalebox{0.9}{
    \begin{tabular}{ll cccc cccc cccc}
        \toprule
        \multirow{2}{*}{\rotatebox{90}{}} &
        \multirow{2}{*}{\textbf{Method}~/~\textbf{Dataset}} & \multicolumn{4}{c}{\textbf{WN18RR}} & \multicolumn{4}{c}{\textbf{NELL-995}} & \multicolumn{4}{c}{\textbf{FB15k-237}}  \\
        \cmidrule(lr){3-6}\cmidrule(lr){7-10}\cmidrule(lr){11-14}
        & & H@1  & H@5 & H@10 & MRR & 
        H@1  & H@5 & H@10 & MRR & 
        H@1  & H@5 & H@10 & MRR
        \\
        \midrule
        \multirow{6}{*}{\rotatebox{90}{\hspace*{-6pt} Interpretable}} &
        MINERVA~\cite{das2017go} & 41.3 & - & 51.3 & 44.8 & 66.3 & - & 83.1 & 72.5 & 
        21.7 & - & 45.6 & 29.3  \\
        &MultiHop (ConvE)~\cite{lin-etal-2018-multi-hop}& 41.4 & 48.1 & 51.7 & 44.8 & 65.6 & - & 84.4 & 72.7 & 32.7 & - & 56.4 & 40.7  \\
        
        & Multihop (ComplEx)~\cite{lin-etal-2018-multi-hop} & 42.5 & 49.4 & 52.6 & 46.1 & 64.4 & 79.1 & 81.6 & 71.2 & \highest{32.9} & - & 54.4 & 39.3 \\
        & AnyBURL (C rules)~\cite{Meilicke2019AnytimeBR}~~ & 42.9 & 51.6 & 53.7 & - & 44.0 & 56.0 & 57.0 & - & 26.9 & 43.1 & 52.0 & - \\
        & RuleGuider (ConvE) & 42.2 & 49.9 & 53.6 & 46.0 & 66.0 & 82.0 & 85.1 & 73.1 & 31.6 & \highest{49.6} & \highest{57.4} & \highest{40.8}  \\ 
        & RuleGuider (ComplEx) & \highest{44.3} & \highest{52.4} & \highest{55.5} & \highest{48.0} & \highest{66.4} & \highest{82.7} & \highest{85.9} & \highest{73.6} & 31.3 & 49.2 & 56.4 & 39.5 \\
        \midrule
        \multirow{4}{*}{\rotatebox{90}{\hspace*{-6pt}Embedding}} 
        & DistMult~\cite{Yang2014EmbeddingEA} & 35.7 & - & 38.4 & 36.7 & 55.2 & - & 78.3 & 64.1 & 32.4 & - & 60.0 & 41.7  \\
        & ComplEx~\cite{Trouillon2016ComplexEF} & 41.5 & 45.6 & 46.9 & 43.4 & 63.9 & 81.7 & 84.8 & 72.1 & 33.7 & 54.0 & \underline{62.4} & 43.2 \\
        & ConvE~\cite{dettmers2018convolutional} & 40.1 & 49.8 & 53.7 & 44.6 & \underline{66.7} & \underline{85.3} & \underline{87.2} & \underline{75.1} & \underline{34.1} & \underline{54.7} & 62.2 & \underline{43.5} \\
        & RotateE~\cite{Sun2019RotatEKG} & \underline{42.2} & \underline{51.3} & \underline{54.1} & \underline{46.4} & - & - & - & - &32.2 & 53.2 & 61.6 & 42.2  \\
        \bottomrule
    \end{tabular}
    }
    \vspace{-.1cm}
    \caption{\textbf{Performance comparison with walk-based approaches}. Best scores among the interpretable methods and embedding-based methods are bold and underlined, respectively. In addition, we present the reported scores for state-of-the-art embedding-based methods as reference. We underscore the best performing ones in this category. 
    }
    \vspace{-0.1cm}
    \label{tb:mainresults}
\end{table*}

\begin{table}[t]
    \centering
    \small
    \scalebox{1.0}{
    \begin{tabular}{l c c c}
    \toprule
    Phase & WN18RR & NELL-995 & FB15K-237\\
    \midrule
    Pre-training & 69.2\% & 44.9\% & 46.1\% \\ 
    Training & 40.7\% & 24.5\% & 41.5\% \\ 
    \bottomrule
    \end{tabular}
    }
    \vspace{-.2cm}
   \caption{Percentage of rules used by RuleGuider (ComplEx) to predict $e_o$ (beam 0) during inference on the development set at the end of pre-training and training phase.
   }
   \vspace{-.2cm}
    \label{tb:rulepercentage}
\end{table}

\section{Experiments}
In this section, we compare RuleGuider with other approaches on three datasets. We describe the experiment setting, results, and analysis.
\subsection{Experimental Setup}

\start{Datasets.} 
We evaluate different methods on three benchmark datasets.
(1) FB15k-237~\cite{toutanova2015representing}, (2) WN18RR~\cite{dettmers2018convolutional}, and (3) NELL-995~\cite{xiong2017deeppath}. 

\start{Hyperparameters.}
We set all embedding size to 200. Each history encoder is a three-layer LSTM with a hidden state dimension 200. 
We use AnyBURL \cite{Meilicke2019AnytimeBR} as the rule miner and set the confidence score threshold to be 0.15. Other hyperparameters are shown in appendix.

\subsection{Results}

Table~\ref{tb:mainresults} shows the evaluation results.
RuleGuider achieves the state-of-the-art results over walk-based methods on WN18RR and NELL-995, and also competitive results on FB15k-237.
One possible reason is: compared to the other two datasets, the relation space in FB15k-237 is much larger and the rules is relatively sparse in the large relational path space, which makes it harder for the relation agent to select a desired rule.

We find symbolic-based models perform strongly on WN18RR and we set $\lambda$ higher to receive stronger rule guidance. We also observe that embedding-based methods have consistently good performance on all datasets compared to walk-based methods despite their simplicity. One possible reason is that embedding-based methods implicitly encode the connectivity of the whole graph into the embedding space \cite{lin-etal-2018-multi-hop}. Embedding-based methods are free from the strict traversing in the graph and sometimes benefit from this property due to incompleteness of the graph. By leveraging rules, we also incorporate some global information as guidance to make up for the potential searching space loss during the discrete inference process.

Table~\ref{tb:rulepercentage} shows the percentage of rules used on the development set using ComplEx embedding in the pre-training and training phase. It shows that our model abandons a few rules to further improve hit performance during the training phase.

\begin{table}[!t]
    \centering
    \small
    \scalebox{1}{
    \begin{tabular}{l c c c c}
    \toprule
    \textbf{Model} & Freeze & No & Single & Ours \\
    \midrule
    H@1 & 41.9& 41.8& 42.4& \highest{42.9}\\
    MRR & 45.5& 45.7& 46.4& \highest{46.5}\\
    \bottomrule
    \end{tabular}
    }
    \vspace{-.2cm}
   \caption{
   \textbf{Freeze}, \textbf{No} and \textbf{Single} represent models with freezing pre-trained relation agent, without pre-training and without separating the agent.}
   \vspace{-.2cm}
    \label{tb:analysis}
\end{table}

\subsection{Ablation Study}
\label{ablation}
We run different variants of RuleGuider on the development set of WN18RR. We use ComplEx hit reward shaping for consistency. Table~\ref{tb:analysis} shows the results. Freezing pre-trained agent performing worse indicates that hit reward is necessary. Removing pre-training performing worse shows that the walk-based agents benefit from logic rules. The single agent variant performing worse shows the effectiveness of pruning action space.

\subsection{Human Evaluation}

\begin{table}[!t]
    \centering
    \small
    \scalebox{1}{
    \begin{tabular}{l c c c c}
    \toprule
     & Multihop  & Tie & RuleGuider \\
    \midrule
    Vote & 36.92\% & 0\% & 63.08\% \\
    \bottomrule
    \end{tabular}
    }
    \vspace{-.2cm}
   \caption{Human evaluation vote between Multihop\cite{lin-etal-2018-multi-hop} and ruleGuider for correctly predicted path on FB15K-237 development set. Both model use ComplEx reward shaping.}
   \vspace{-.2cm}
    \label{tb:humanEval}
\end{table}

Besides the evaluation metrics, we further analyze whether the reasoning path that leads to correctly predicted entity is reasonable. We perform human evaluation on the Amazon Mechanical Turk. We randomly sample a evaluation set with 300 triples from development set using uniform distribution on FB15k-237.  During evaluation, given the ground truth triple, three judges are asked to choose which path is a better explanation/decomposition of it between: 1. path generated by our method. 2. paths generated by Multihop's method. 3. Draw or none of them are reasonable. Note that there are 2.6\% of the predicted paths are the same and they are excluded from the evaluation set. For each triple, we count majority vote as the evaluation result. As it's possible that the three judges each choose a different option which leads to one vote on each choice. In this case, we do not count it in the final evaluation result (table~\ref{tb:humanEval}). Unexpectedly, no triple get more than one vote on Tie. RuleGuider achieves a better performance and the reasoning path makes more sense to human judges comparing to Multihop with ComplEx reward shaping.






%% file: src/6-Conclusion.tex
\section{Conclusions}
In this paper, we proposed a collaborative framework utilizing both symbolic-based and walk-based models. We separate the walk-based agent into an entity and relation agent to effectively leverage the symbolic rules and significantly reduce the action space.
Experimentally, our approach improved the performance of the state-of-the-art walk-based models on two benchmark KGs.


In future work, we would like to study how to introduce acyclic rules to the walk-based systems.

%% file: appendix.tex
\appendix
\section{Appendix}
\subsection{Datasets}
We use the same training, development and testing set splits as \citet{lin-etal-2018-multi-hop}. Following \citet{lin-etal-2018-multi-hop}, we restrict the output degree of an entity by selecting top $\eta$ neighbors according to their PageRank score~\cite{page1999pagerank}. We remove unseen entities in test set for NELL-995. We also add reverse links from object entity $e_o$ to subject entity $e_s$.

The detailed datasets statistics is shown in Table~\ref{tb:data} and Table~\ref{tb:kgstats}.
\begin{table}[h]
    \small
    \centering
    \setlength\tabcolsep{4pt}
    \begin{tabular}{lrrr}
        \hline
        \textbf{Dataset} & \textbf{\#Ent} & \textbf{\#Rel} & \textbf{\#Fact} \\
        \hline 
        FB15k-237 & 14,505 & 237 & 272,115\\
        WN18RR & 40,945 & 11 & 86,835 \\
        NELL-995 & 75,492 & 200 & 154,213 \\
        \hline
    \end{tabular}
    \caption{Number of relations, entities and fact triples on three datasets.}
    \label{tb:data}
\end{table}

\begin{table}[h]
    \centering
    \small
    \scalebox{0.8}{
    \begin{tabular}{l cc cc cc}
    \toprule
        \multirow{2}{*}{\textbf{Dataset}} & \multicolumn{2}{c}{\textbf{Degree}} & \multicolumn{2}{c}{\textbf{Relation Degree}} & \multicolumn{2}{c}{\textbf{Entity Degree}}  \\
        \cmidrule(lr){2-3}\cmidrule(lr){4-5}\cmidrule(lr){6-7}
        & mean  & median & 
         mean  & median & 
         mean  & median
        \\
    \midrule
    FB15k-237 & 37.52 & 22 & 10.32 & 10 & 29.17 & 18\\
    NELL-995  & 4.03 & 1 & 1.79  & 1 & 3.47 & 1\\ 
    WN18RR & 4.28 & 3 & 2.55 & 2 & 3.54 & 2\\ 
    \bottomrule
    \end{tabular}
    }
    \vspace{-.2cm}
  \caption{Output degree of each entity on three datasets. Degree is the total edges incident to each entity. Relation degree is the number of relations on each entity's output edges. Entity degree is the number of entities that each entity connects to.}
  \vspace{-.2cm}
    \label{tb:kgstats}
\end{table}

\subsection{Training Details}
\subsubsection{Hardware and Runtime}
We trained our model on one NVIDIA GeForce 1080 Ti GPU. Table~\ref{tb:runtime} shows the runtime detail of our model training.

\begin{table}[h]
    \small
    \centering
    \scalebox{0.8}{
    \begin{tabular}{lrrr}
        \hline
        \textbf{} & \textbf{WN18RR} & \textbf{NELL-995} & \textbf{FB15k-237} \\
        \hline 
        number of epochs & 50 & 1000 & 30 \\
        time/epoch & 360s & 50s & 1800s \\
        trainable parameters & 26M & 47M & 11M \\
        \hline
    \end{tabular}}
    \caption{Running time and model parameters.}
    \label{tb:runtime}
\end{table}

\subsubsection{Hyperparamters Search}
We use Adam to train our model and use beam search during inference to give the ranked prediction of object entities. We run grid search on a bunch of hyperparameters to select the best configuration. The bounds for searched hyperparameters are shown in Table~\ref{tb:gridsearch}. An exception is rule guidance reward ratio $\lambda$, which are manually tuned. Table~\ref{tb:hyperpara} shows the configurations of our best performing model.
\begin{table}[h]
    \small
    \centering{
    \begin{tabular}{lr}
        \hline
        \textbf{Hyperparameter} & \textbf{Search Bounds} \\
        \hline 
        regularization weight $\beta$ & [0.0, 0.1] \\
        embedding dropout rate & [0.0, 0.3] \\
        hidden layer dropout rate & [0.0, 0.3] \\
        relation dropout rate & [0.0, 0.95] \\
        entity dropout rate & [0.0, 0.95] \\
        bandwidth & \{200, 256, 400, 512\} \\
        mini-batch size & \{64, 128, 256, 512\} \\
        learning rate & [0.001, 0.003] \\
        number of hops & \{2, 3\} \\
        \hline
    \end{tabular}}
    \caption{Searched hyperparameters using grid search. Following~\cite{lin-etal-2018-multi-hop}, we add an entropy regularization term in the training objective and the term is weighted by regularization weight $\beta$. Bandwidth is the entity output degree.}
    \label{tb:gridsearch}
\end{table}

\subsubsection{Confidence Score Threshold}
We analyze the performance of our model with different confidence score thresholds of the rule miner (AnyBURL).
We set the maximum threshold to be 0.15.
Table~\ref{tb:threshold} shows the results. 
The results do not present any observable pattern. One potential reason is that walk-based reasoning paths and less confident rules may have similar performance on certain queries.

\begin{table}[h]
    \centering
    \small
    \scalebox{1}{
    \begin{tabular}{l c c c c}
    \toprule
    Confidence & $> 0.00$ & $> 0.05$ & $> 0.10$ & $> 0.15$\\
    \midrule
    @1 & 42.7 & 42.7& 42.4& \textbf{42.9}\\
    MRR & 46.3& \textbf{46.6}& 46.0& 46.5\\
    \bottomrule
    \end{tabular}
    }
    \vspace{-.2cm}
   \caption{Different confidence score thresholds.
   }
   \vspace{-.2cm}
    \label{tb:threshold}
\end{table}

\begin{table}[h]
    \small
    \centering
    \scalebox{0.8}{
    \begin{tabular}{lrrr}
        \hline
        \textbf{Hyperparameter} & \textbf{WN18RR} & \textbf{NELL-995} & \textbf{FB15k-237} \\
        \hline 
        regularization weight $\beta$ & 0.0 & 0.05 & 0.02 \\
        embedding dropout rate & 0.1 & 0.1 & 0.3 \\
        hidden layer dropout rate & 0.1 & 0.1 & 0.1 \\
        relation dropout rate & 0.1 & 0.1 & 0.5 \\
        entity dropout rate & 0.1 & 0.1 & 0.5 \\
        bandwidth & 500 & 256 & 400 \\
        mini-batch size & 256 & 128 & 256\\
        learning rate & 0.001 & 0.001 & 0.0015 \\
        number of hops & 3 & 3 & 3 \\
        rule guidance reward ratio $\lambda$ & 0.65 & 0.1 & 0.1 \\
        \hline
    \end{tabular}}
    \caption{Hyperparameter used in our model.}
    \label{tb:hyperpara}
\end{table}

\subsubsection{Evaluation Metrics}
During inference, the model gives a ranked list of predicted entities as the result. We use Hit@N (H@N) and Mean Reciprocal Rank (MRR) to evaluate the model performance based on the ranked list. Hit@N measures the percentage of test triples for which the correct object entity is ranked in top N in the candidate entities list. MRR measures the average reciprocal of the rank of the object entity.

\subsection{Development Set Performance}
The result of RuleGuilder with complex embedding hit reward shaping on development set is shown in Table~\ref{tb:devresults}.
\begin{table}[h]
    \centering
    \small
    \scalebox{1}{
    \begin{tabular}{l cccc cccc cccc}
        \toprule
        \textbf{Dataset} &  H@1  & H@5 & H@10 & MRR \\
        \midrule
        WN18RR & 43.7 & 50.2 & 53.0 & 47.0 \\
        \midrule
        NELL-995 & 74.3 & 92.9 & 93.4 & 83.0 \\
        \midrule
        FB15K-237 &  27.4 & 47.4 & 55.8 & 36.7  \\
        \bottomrule
    \end{tabular}
    }
    \vspace{-.2cm}
    \caption{Performance of RuleGuider using ComplEx embedding on development set.}
    \vspace{-.3cm}
    \label{tb:devresults}
\end{table}